\documentclass{article}
\usepackage{arxiv}
\usepackage[utf8]{inputenc} 
\usepackage[T1]{fontenc}    
\usepackage{hyperref}       
\usepackage{url}            
\usepackage{booktabs}       
\usepackage{amsfonts}       
\usepackage{nicefrac}       
\usepackage{microtype}      
\usepackage{lipsum}		
\usepackage{multirow} 
\usepackage{amsmath} 
\usepackage{graphicx}
\usepackage[square, numbers]{natbib}
\usepackage{doi}
\usepackage{subcaption}
\usepackage{float}
\usepackage{tabularx} 

\title{PHYSICS-INFORMED NEURAL NETWORKS FOR MODELING OCEAN POLLUTANT}

\date{April 24, 2025}	

\author{ \href{https://orcid.org/0009-0000-2783-6711}{\includegraphics[scale=0.06]{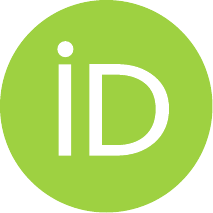}\hspace{1mm}Karishma Battina}\\
	Indiana Wesleyan University\\
	Indianapolis, IN 46260 \\
	\texttt{karishma.battina@myemail.indwes.edu} \\
     \and
    \href{https://orcid.org/0009-0008-5476-9962}{\includegraphics[scale=0.06]{orcid.pdf}\hspace{1mm}Prathamesh Dinesh Joshi} \\
    Vizuara AI Labs \\
    Pune, India \\
    \texttt{prathamesh@vizuara.com} \\
    \and
    {\hspace{1mm}Raj Abhijit Dandekar} \\
    Vizuara AI Labs \\
    Pune, India \\
    \texttt{raj@vizuara.com} \\
    \and
    {\hspace{1mm}Rajat Dandekar} \\
    Vizuara AI Labs \\
    Pune, India \\
    \texttt{rajatdandekar@vizuara.com} \\
    \and
    {
    \hspace{1mm}Sreedath Panat} \\
    Vizuara AI Labs \\
    Pune, India \\
    \texttt{sreedath@vizuara.com} \\
}
        
\renewcommand{\shorttitle}{\textit{arXiv} Template}

\hypersetup{
pdftitle={A template for the arxiv style},
pdfsubject={q-bio.NC, q-bio.QM},
pdfauthor={David S.~Hippocampus, Elias D.~Striatum},
pdfkeywords={First keyword, Second keyword, More},
}

\begin{document}
\maketitle

\begin{abstract}
    Traditional numerical methods often struggle with the complexity and scale of modeling pollutant transport across vast and dynamic oceanic domains. This paper introduces a Physics-Informed Neural Network (PINN) framework to simulate the dispersion of pollutants governed by the 2D advection-diffusion equation. The model achieves physically consistent predictions by embedding physical laws and fitting to noisy synthetic data, generated via a finite difference method (FDM), directly into the neural network training process. This approach addresses challenges such as non-linear dynamics and the enforcement of boundary and initial conditions. Synthetic data sets, augmented with varying noise levels, are used to capture real-world variability. The training incorporates a hybrid loss function including PDE residuals, boundary/initial condition conformity, and a weighted data fit term. The approach takes advantage of Julia's scientific computing ecosystem for high-performance simulations, offering a scalable and flexible alternative to traditional solvers.
\end{abstract}


\section{Introduction}


The dispersion of pollutants in ocean currents, from plastic debris to oil spills, poses significant threats to marine ecosystems and coastal communities. Predicting pollutant transport requires solving the advection-diffusion equation, a partial differential equation(PDE) that models the interplay of fluid-driven advection and turbulent diffusion \cite{Fischer1979}. However, traditional numerical methods, such as finite-volume or spectral approaches, face critical limitations: computational costs escalate exponentially with domain size, parameter uncertainties degrade accuracy, and resolving boundary effects near coastlines remains challenging \cite{LeVeque2002, Chassignet2006, Morton2005}. These constraints hinder timely predictions for environmental mitigation efforts, particularly in dynamic, large-scale oceanic systems.

Recent advances in scientific machine learning have introduced physics-informed neural networks (PINNs) as a paradigm for solving PDEs \cite{Reichstein2019}. Unlike conventional solvers, PINNs embed physical laws directly into neural network training, enabling mesh-free simulations and seamless integration of sparse observational or synthetic data. While PINNs have shown promise in fluid dynamics, their application to ocean pollution remains underexplored, especially in addressing real-world complexities such as transient velocity fields, spatially varying diffusion, and irregular coastal geometries.

In this work, we develop a PINN-based framework tailored for oceanic pollutant transport. Our key contributions are threefold:

\subsection{Review of Pollutant Transport Modeling}
The modeling of oceanic pollutant transport has traditionally been approached from two main perspectives. Eulerian models, such as the Finite Difference Method (FDM) or Finite Element Method (FEM), discretize the ocean into a fixed grid and solve the governing equations at each grid point \cite{Smith1983}. These methods are robust but become computationally prohibitive for large-scale, long-term simulations. Alternatively, Lagrangian particle-tracking models simulate the trajectories of individual pollutant particles, which is effective for tracking sparse debris but can struggle to represent continuous concentration fields accurately \cite{Zhang2008}.

Recent years have seen the application of machine learning to augment or replace these methods. Machine learning techniques augment traditional models but face generalization challenges without physical consistency \cite{Reichstein2019, Lguensat2022}.
Some studies use data-driven neural networks to predict flow fields or correct biases in numerical models \cite{Zhu2019}. However, these purely data-driven approaches require vast amounts of training data and may not generalize well to scenarios outside their training distribution, as they lack inherent knowledge of the underlying physics \cite{Zhu2018}.

\subsection{Physics-Informed Neural Networks (PINNs)}
To address these limitations, recent advances in scientific machine learning have introduced Physics‐Informed Neural Networks (PINNs) as a paradigm for solving PDEs \cite{17physics, Cuomo2022, Karniadakis2021}. Unlike conventional solvers or purely data‐driven models, PINNs embed the physical laws directly into the neural network’s loss function \cite{Shankar2022}. This enables mesh‐free simulations and seamless integration of sparse observational or synthetic data \cite{Zhu2019, Sun2020}. While PINNs have shown promise in fluid dynamics \cite{Sun2020}, their application to ocean pollution remains underexplored, especially concerning the systematic analysis of how network design choices impact solution accuracy for complex, noisy systems \cite{wang2022when, saadat2022neural, paszynski2023physics}.

In this work, we develop a PINN‐based framework tailored for oceanic pollutant transport. Our key contributions are:
\begin{itemize}
  \item The development of a robust, data‐augmented PINN framework in Julia that accurately enforces sharp initial conditions through a hybrid, weighted loss function.
  \item A systematic investigation into the role of neural network architecture (depth, width) and hyperparameters (learning rate, optimizers) on solution accuracy and computational cost.
  \item A comprehensive performance benchmark that identifies an optimal configuration capable of achieving a relative $L_2$ error of approximately 8.25\% against a high‐resolution FDM.
\end{itemize}




\section{Methodology}
\label{sec:others}

\subsection{Mathematical Formulation of the Advection-Diffusion Equation}
The advection-diffusion equation governs the transport of pollutants in oceanic environments, a partial differential equation that captures the combined effects of advection (directional transport by currents) and diffusion (random spreading of particles) \cite{Okubo1971, Fischer1979}:

\begin{equation}
    \frac{\partial u}{\partial t} + \mathbf{v} \cdot \nabla u = D \nabla^2 u
\end{equation}

\noindent\textbf{Where:}
\begin{description}
  \item[$u(x,y,t)$] Represents the concentration of pollutants at spatial location $(x,y)$ and time $t$.
  \item[$\mathbf{v}=(v_x,v_y)$] is the velocity field representing ocean currents, with components $v_x$ and $v_y$. In this study, we set:
    \[
      v_x = 0.5,\quad v_y = 0.5.
    \]
  \item[$D$] is the diffusion coefficient, quantifying the rate of pollutant dispersion \cite{lu2021deepxde, Bezanson2017}. For this work, we use
    \[
      D = 0.01.
    \]
\end{description}

The advection term \begin{equation}
    \mathbf{v} \cdot \nabla u \quad\end{equation} models the transport of pollutants by ocean currents, while the diffusion term \begin{equation} \quad D \nabla^2 u
\end{equation} accounts for the spread of pollutants due to random molecular motion and turbulent eddies. When expanded in two dimensions, this equation becomes the following:

\begin{equation}
    \frac{\partial u}{\partial t} + v_x \frac{\partial u}{\partial x} + v_y \frac{\partial u}{\partial y} = D \left( \frac{\partial^2 u}{\partial x^2} + \frac{\partial^2 u}{\partial y^2} \right)
\end{equation}

This formulation serves as the foundation for our PINN-based modeling approach, enabling the simulation of pollutant transport across oceanic domains with varying current patterns and diffusion characteristics.

\subsection{Physics-Informed Neural Network Framework}
Our approach leverages a PINN to approximate the solution u(t,x,y) of the advection-diffusion equation. The framework is designed to be expressive, physically consistent, and robust to noisy data through a hybrid training strategy.

\subsubsection{Boundary and Initial Conditions}

To ensure a well‐posed problem, we define the spatio‐temporal domain as
\[
  t \in [0,\,0.25], \quad x \in [0,\,1.0], \quad y \in [0,\,1.0].
\]
The final time is set to 
\[
  T_{\mathrm{final}} = 0.25
\]
to ensure the reference solution remains non‐trivial within the domain.

The initial condition represents a sharply localized pollutant release at the center of the domain. This Gaussian peak is a standard analytical choice for modeling an idealized, instantaneous point-source spill before significant dispersion occurs \cite{fu2025long, Karniadakis2021}.

Dirichlet boundary conditions are imposed, enforcing approximately zero pollutant concentration at the domain edges. This setup represents a localized pollution event within a much larger body of water, where the concentration at the far-field boundaries is assumed to be negligible. The inclusion of stochastic noise in these conditions aims to simulate the minor fluctuations and measurement uncertainties present in real-world oceanic systems.

The following boundary and initial conditions, incorporating stochastic noise to simulate realistic data, are imposed:

\textbf{Boundary Conditions (Dirichlet):}\newline
\begin{equation}u(t,\,x,\,0) \approx 0.0 + \epsilon_{\mathrm{BC}}\end{equation}
\begin{equation}u(t,\,x,\,1) \approx 0.0 + \epsilon_{\mathrm{BC}}\end{equation}
\begin{equation}u(t,\,0,\,y) \approx 0.0 + \epsilon_{\mathrm{BC}}\end{equation}
\begin{equation}u(t,\,1,\,y) \approx 0.0 + \epsilon_{\mathrm{BC}}\end{equation}

These conditions enforce approximately zero pollutant concentration at the domain boundaries \cite{Zhang2008}.\newline
\( \epsilon_{\mathrm{BC}} \) represents additive Gaussian noise with a standard deviation of \( \sigma_{\mathrm{BC}} = 0.01 \).

\textbf{Initial condition (IC):}
\[
u(0, x, y) \approx \exp\left(-100\left[(x - 0.5)^2 + (y - 0.5)^2\right]\right) \cdot \left(1 + \epsilon_{\mathrm{IC}}\right)
\]

This initial condition represents a sharply localized pollutant release at the center of the domain. \newline
\(\epsilon_{\mathrm{IC}}\) is Gaussian noise with a standard deviation of 0.5\% of the clean initial condition value.

\subsubsection{Neural Network Architecture}
The core of the Physics-Informed Neural Network (PINN) is a neural network that approximates the solution \( u(t, x, y) \) of the partial differential equation of advection-diffusion. The chosen architecture is a fully connected feedforward neural network, designed to be sufficiently expressive to capture the complex spatio-temporal dynamics of pollutant dispersion.

The network architecture comprises:

\begin{itemize}
    \item \textbf{Input Layer:} Receives 3 input features: the temporal coordinate \( t \) and the spatial coordinates \( (x, y) \).
    
    \item \textbf{Hidden Layers:} The number of layers and neurons were varied. The primary architectures tested were:
    \begin{itemize}
  \item 9 hidden layers with 64 neurons per layer
  \item 9 hidden layers with 128 neurons per layer
  \item 9 hidden layers with 256 neurons per layer
\end{itemize}
    
    \item \textbf{Activation Function:} The hyperbolic tangent (\texttt{Tanh}) activation function is applied to the output of each neuron in all hidden layers. \texttt{Tanh} is chosen for its smoothness and non-zero second derivatives, which are beneficial for computing PDE residuals via automatic differentiation.
    \item \textbf{Output Layer:} A single neuron with a linear activation function produces the scalar value \( \hat{u}(t, x, y) \), representing the predicted pollutant concentration.
\end{itemize}

The parameters (weights and biases) of the neural network are initialized using default schemes in \texttt{Lux.jl} library and all network operations are performed using Float32 precision for computational efficiency \cite{Bezanson2017, Rackauckas2017}.

\subsubsection{Synthetic Data and Noise Augmentation}

Synthetic training data is generated by numerically solving the advection-diffusion equation using a Finite Difference Method (FDM). The FDM solver operates on a discrete grid with \( N_x = 51 \) points in the \( x \)-direction, \( N_y = 51 \) points in the \( y \)-direction, and \( N_t = 100 \) time steps, discretizing the domain up to \( T_{\text{final}} = 0.25 \) yielding a total of 262{,}701 \text{ data points}.

To enhance model robustness and mimic real-world measurement uncertainties, Gaussian noise is added to the clean FDM solution \( U_{\text{clean}} \):
\[
U_{\text{noisy}} = U_{\text{clean}} + \varepsilon_{\text{data}}
\]
where \( \varepsilon_{\text{data}} \) is drawn from a normal distribution with a standard deviation of\[
\sigma_{\text{data}} = 0.005 \cdot \text{std}(U_{\text{clean}})
\]

\subsubsection{Training Procedure}
The PINN is trained by finding the optimal set of network parameters \( \theta \) that minimize a composite loss function. This loss function ensures the network's output  
\( \hat{u}(t, x, y; \theta) \) both fits the available data and adheres to the governing physical laws.

\subsection*{Loss Function Formulation}

The total loss function \( \mathcal{L}_{\text{total}}(\theta) \) is a sum of three main components: a physics-based loss, a specific initial condition loss, and a data fidelity loss:
\[
\mathcal{L}_{\text{total}}(\theta) = \mathcal{L}_{\text{physics}}(\theta) + w_{\text{IC}}\mathcal{L}_{\text{IC}}(\theta) + w_{\text{data}} \mathcal{L}_{\text{data}}(\theta)
\]

Each component is defined as follows:
\paragraph{Physics-Based Loss ($L_{\mathrm{physics}}$)}  
This term is automatically constructed by \texttt{NeuralPDE.jl} \cite{Rackauckas2020}. It includes the mean squared error (MSE) of the PDE residual, the symbolic boundary condition residuals, and the symbolic initial condition residual, all evaluated over a set of collocation points. For this study, 200 collocation points were sampled per batch using a Latin Hypercube algorithm.

\paragraph{Initial Condition Loss ($L_{\mathrm{ic}}$)}  
To ensure the network precisely learns the initial state, an explicit, heavily‐weighted data loss term is added for the initial condition. This term computes the MSE between the PINN’s predictions at $t=0$ and the clean initial condition data, weighted by 
\[
  w_{\mathrm{ic}} = 500.0\,.
\]

\paragraph{Data Fidelity Loss ($L_{\mathrm{data}}$)}  
This term encourages the network to fit the noisy FDM data over the entire spatio‐temporal domain. It is defined as the MSE between the PINN’s predictions and the noisy FDM data, weighted by 
\[
  w_{\mathrm{data}} = 10.0\,.
\]

Both $L_{\mathrm{ic}}$ and $L_{\mathrm{data}}$ are computed using mini‐batches of size 1,024, randomly sampled at each training iteration.



\subsection*{Optimization Strategy}

Our optimization strategy involved a two-stage approach to robustly minimize the total loss. The primary optimizers explored were ADAM \cite{Kingma2014} and its variant ADAMW \cite{Loshchilov2017}, known for their effectiveness in training deep neural networks \cite{Schmidt2020}. The performance of these optimizers was systematically investigated by varying key hyperparameters, including the learning rate and the total number of training iterations. Additionally, a quasi-Newton method (L-BFGS) \cite{Nocedal1980} was evaluated as a second-stage refiner, initialized with the parameters obtained from the Adam-family optimizers, to fine-tune the solution. This two-step approach of using Adam followed by L-BFGS is a common strategy in PINN literature \cite{Jarlebring2022}, with other hybrid methods that combine global search algorithms like CMA-ES with gradient descent also proving effective \cite{yang2024effective}. This multi-faceted approach allowed for a thorough exploration of the optimization landscape to identify the most effective configuration for this problem.

\section{Results}
This section presents a comprehensive evaluation of the Physics-Informed Neural Network (PINN) framework applied to the 2D advection-diffusion equation. The analysis emphasizes accuracy, computational efficiency, and sensitivity to various hyperparameters such as neural network architecture, optimizer choices, learning rates, and training duration. All experiments used a final simulation time of $T_{\text{final}} = 0.25$ and a $51\times 51$ Finite Difference Method (FDM) grid to generate reference data.

\subsection{Hyperparameter Configuration}
The experimental design included fixed physical and numerical parameters while varying key neural network settings and optimization. Table~\ref{tab:hyperparameters}  summarizes these configurations.

\begin{table}[h!]
\centering
\caption{Summary of Fixed and Varied Hyperparameters}
\label{tab:hyperparameters}
\begin{tabularx}{\textwidth}{@{}l l X@{}}
\toprule
\textbf{Parameter Category} & \textbf{Parameter(s)} & \textbf{Value(s) / Setting(s)} \\
\midrule
\multirow{3}{*}{Physical Parameters} 
 & Final Time ($T_{\text{final}}$) & 0.25 \\
 & Domain Max ($x_{\text{max}}, y_{\text{max}}$) & 1.0, 1.0 \\
 & Diffusion Coefficient ($D$) & 0.01 \\
 & Advection Velocity ($v_x, v_y$) & 0.5, 0.5 \\
\midrule
\multirow{2}{*}{Numerical Parameters (FDM)}
 & Grid Size ($N_x, N_y$) & 51, 51 \\
 & Time Steps ($\Delta t$) & 100 \\
\midrule
\multirow{3}{*}{Noise Levels}
 & Symbolic BC Absolute Noise & 0.01 \\
 & Symbolic IC Relative Noise & 0.005 \\
 & FDM Data Relative Noise & 0.005 \\
\midrule
\multirow{3}{*}{Loss Function Setup}
 & Data Loss Weight ($W_{\text{data}}$) & 10.0 \\
 & IC Loss Weight ($W_{\text{IC}}$) & 500.0 \\
 & Data Batch Size & 1024 \\
 & IC Batch Size & 1024 \\
\midrule
\multirow{1}{*}{Activation Function (NN)}
 & Activation Function (used consistently) & Tanh \\
\midrule
\multirow{1}{*}{Varied Hyperparameters}
 & Neural Network Architecture & 9L-64N, 9L-128N, 9L-256N \\
  & Optimizers & ADAM, ADAM+LBFGS, ADAMW \\
    & Learning Rates (Adam/AdamW) & 0.002, 0.005, 0.006, 0.007, 0.008, 0.01 \\
    & Iterations (Adam/AdamW) & 1000, 2000, 3000, 6000, 10000 \\
    & LBFGS Iterations & Typically 500 (used post-Adam) \\
\bottomrule
\end{tabularx}
\end{table}


\subsection{Experimental Scenarios and Performance}
Different scenarios were executed, exploring different combinations of network architectures, optimizers, learning rates, and iteration counts. The key performance metrics – final selected loss, relative L2 error at Tfinal=0.25, and PINN training execution time – are summarized in Table 2. The FDM solver consistently took approximately 0.25-0.28 seconds to generate the 51×51×101 reference dataset. PINN inference times for full-field predictions were also consistently fast, around 0.024-0.029 seconds.

\begin{table}[H]
  \centering
  \caption{Performance Summary of PINN Training Scenarios}
  \label{tab:pinn_performance}
  \begin{tabular}{|l|l|r|r|r|r|r|}
    \hline
    \textbf{NN Arch.} & \textbf{Optimizer}  & \textbf{Iterations} & \textbf{LR} & \textbf{Final Loss}     & \textbf{Rel. $L_2$ Error} & \textbf{Train Time(s)} \\
    \hline
    9L, 64N   & ADAM+LBFGS & 1,000  & 0.002 & $4.0803\times10^{1}$ & 0.8278  & 84.45   \\
    9L, 64N   & ADAM+LBFGS & 3,000  & 0.002 & $1.0340\times10^{1}$ & 0.41047 & 321.32  \\
    9L, 64N   & ADAM       & 10,000 & 0.002 & $2.6683\times10^{0}$ & 0.18720 & 597.99  \\
    9L, 64N   & ADAMW      & 10,000 & 0.002 & $2.6683\times10^{0}$ & 0.18720 & 601.42  \\
    9L, 128N  & ADAMW      & 6,000  & 0.002 & $3.2537\times10^{0}$ & 0.17700 & 933.14  \\
    9L, 128N  & ADAM       & 6,000  & 0.005 & $2.5142\times10^{0}$ & 0.13490 & 942.05  \\
    9L, 128N  & ADAM       & 6,000  & 0.002 & $3.2537\times10^{0}$ & 0.17702 & 1207.19 \\
    9L, 128N  & ADAM       & 6,000  & 0.006 & $3.9296\times10^{0}$ & 0.08245 & 1168.53 \\
    9L, 128N  & ADAM       & 10,000 & 0.006 & $1.2614\times10^{0}$ & 0.12940 & 1987.39 \\
    9L, 256N  & ADAM       & 6,000  & 0.006 & $5.3366\times10^{0}$ & 0.22307 & 3255.62 \\
    \hline
  \end{tabular}
\end{table}

\subsection{Analysis of Results}

\subsubsection{Initial Condition Learning}

Across all documented scenarios, the PINN demonstrated excellent capability in learning the initial condition when using the \texttt{combined\_additional\_loss} strategy with a strong \texttt{ic\_loss\_weight} of 500.0. As consistently shown in the top‐right panel of the four‐panel solution plots (e.g., “PINN IC ($t=0$, ICW:500.0)” in Scenario 1 (Figure 1)), the PINN’s reconstruction of the initial Gaussian peak at $t=0$ almost perfectly matched the “Clean IC ($t=0$)” (top‐left panel). This indicates the effectiveness of the specific IC loss term in anchoring the solution at its starting state.

\subsubsection{Performance with “9 Layers, 64 Neurons” Architecture (Scenarios 1–8)}

\paragraph{ADAM+LBFGS (Scenarios 1, 2)} These scenarios combined ADAM pre‐training with a subsequent L‐BFGS refinement. In Scenario 1 (1,000 ADAM iterations), the model exhibited a high final loss of $4.08\times10^{1}$ and relative $L_2$ error of 0.8278. Increasing to 3,000 ADAM iterations in Scenario 2 reduced the loss to $1.03\times10^{1}$ and the $L_2$ error to 0.41047. In both cases, L‐BFGS did not substantially improve upon the ADAM‐only results. The solution plots (e.g., Scenario 1, Figure 1; Scenario 2, Figure 1) demonstrate accurate initial‐condition learning but reveal artifacts and deviations in the PINN prediction at $T_{\mathrm{final}}$ when compared with the FDM reference.

\begin{figure}[H]
  \centering
  \begin{subfigure}[b]{0.32\textwidth}
    \centering
    \includegraphics[width=\textwidth]{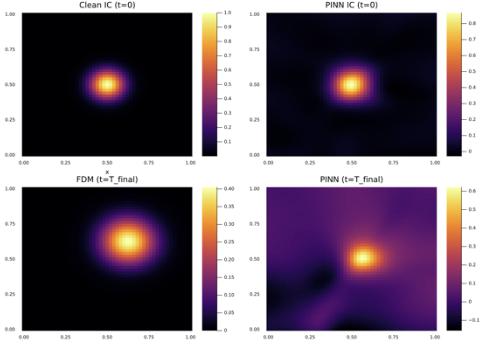}
    \caption{Solution plot}
    \label{fig:three:a}
  \end{subfigure}
  \hfill
  \begin{subfigure}[b]{0.32\textwidth}
    \centering
    \includegraphics[width=\textwidth]{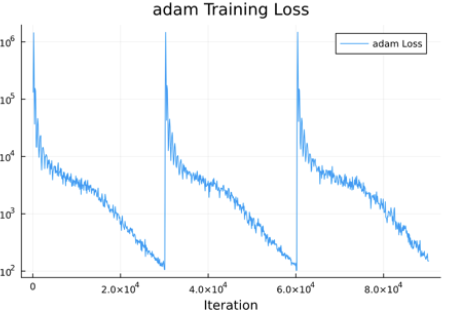}
    \caption{Adam Loss}
    \label{fig:three:b}
  \end{subfigure}
  \hfill
  \begin{subfigure}[b]{0.32\textwidth}
    \centering
    \includegraphics[width=\textwidth]{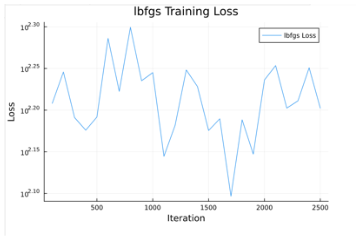}
    \caption{LBFGS Loss}
    \label{fig:three:c}
  \end{subfigure}
  \caption{Solution plots and Adam loss history for Scenario 1 (9L-64N network, 1000 iterations)}
  \label{fig:scenario1-adam-lbfgs}
\end{figure}

\begin{figure}[H]
  \centering
  \begin{subfigure}[b]{0.32\textwidth}
    \centering
    \includegraphics[width=\textwidth]{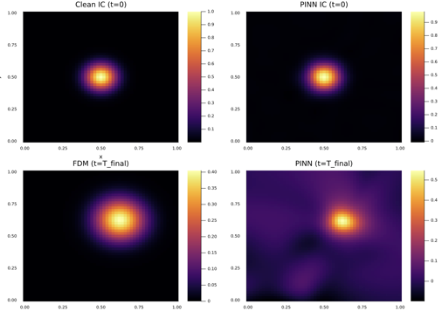}
    \caption{Solution plot}
    \label{fig:three:a}
  \end{subfigure}
  \hfill
  \begin{subfigure}[b]{0.32\textwidth}
    \centering
    \includegraphics[width=\textwidth]{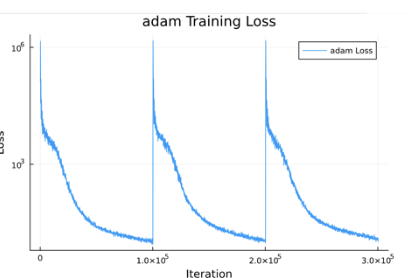}
    \caption{Adam Loss}
    \label{fig:three:b}
  \end{subfigure}
  \hfill
  \begin{subfigure}[b]{0.32\textwidth}
    \centering
    \includegraphics[width=\textwidth]{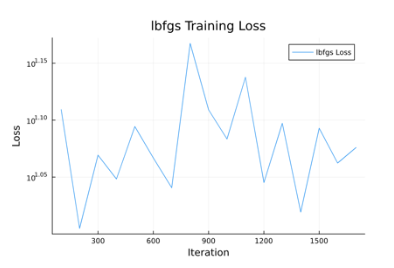}
    \caption{LBFGS Loss}
    \label{fig:three:c}
  \end{subfigure}
  \caption{Solution plots and Adam and LBFGS loss history for Scenario 2 (9L-64N network, 3000 iterations)}
  \label{fig:scenario2-adam-lbfgs}
\end{figure}

\paragraph{ADAM Only:} Performance improved with additional ADAM iterations. The increase in iterations to 2,000 achieved a final loss of $1.44\times10^{1}$ and a relative $L_2$ error of 0.475. The increase in iterations to 6,000 further reduced the loss to $4.08\times10^{0}$ with an $L_2$ error of 0.2776. The best results for this architecture were obtained in Scenario 3 (10,000 iterations), resulting in a loss of $2.67\times10^{0}$ and an $L_2$ error of 0.1872.

\begin{figure}[H]
  \centering
  \begin{subfigure}[b]{0.45\textwidth}
    \centering
    \includegraphics[width=\textwidth]{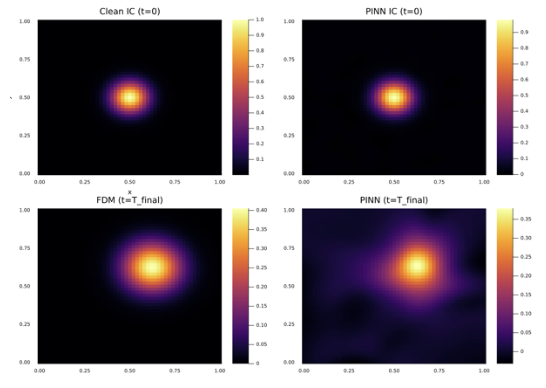}
    \caption{Caption 1}
    \label{fig:three:a}
  \end{subfigure}
  \hfill
  \begin{subfigure}[b]{0.45\textwidth}
    \centering
    \includegraphics[width=\textwidth]{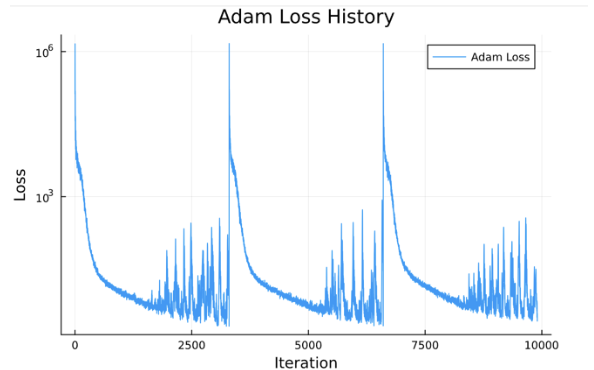}
    \caption{Caption 2}
    \label{fig:three:b}
  \end{subfigure}
  \caption{Solution plots and Adam loss history for Scenario 3 (9L-64N network, 10,000 Adam iterations)}
  \label{fig:scenario2-adam-lbfgs}
\end{figure}

\paragraph{ADAMW Only:} The ADAMW optimizer produced results nearly identical to ADAM for equivalent iteration counts and learning rates. In Scenario 4 (10,000 iterations), ADAMW exactly matched the performance of Scenario 3 (final loss $2.67\times10^{0}$, $L_2$ error 0.1872), and training times were comparable between the two optimizers.

\begin{figure}[H]
  \centering
  \begin{subfigure}[b]{0.45\textwidth}
    \centering
    \includegraphics[width=\textwidth]{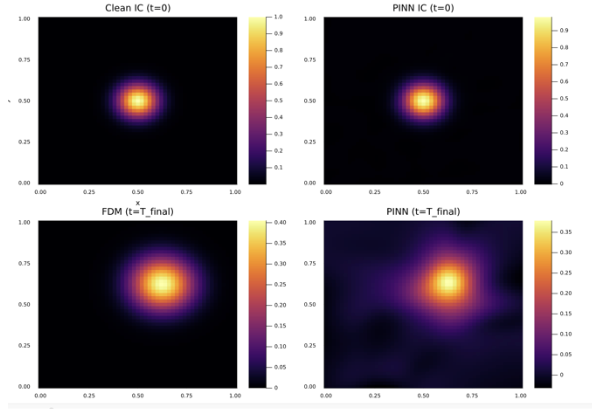}
    \caption{Caption 1}
    \label{fig:three:a}
  \end{subfigure}
  \hfill
  \begin{subfigure}[b]{0.45\textwidth}
    \centering
    \includegraphics[width=\textwidth]{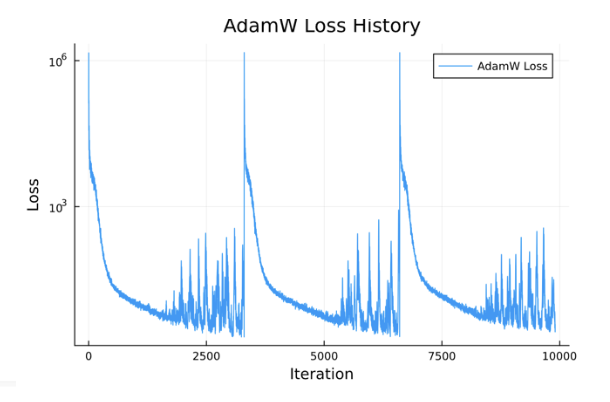}
    \caption{Caption 2}
    \label{fig:three:b}
  \end{subfigure}
  \caption{Solution plots and AdamW loss history for Scenario 4 (9L-64N network, 10,000 Adam iterations)}
  \label{fig:scenario2-adam-lbfgs}
\end{figure}

\subsubsection{Performance with “9 Layers, 128 Neurons” Architecture}

This larger architecture demonstrated the potential for significantly improved accuracy.
\paragraph{ADAMW (Scenario 5)}  
With a learning rate of 0.002 and 6,000 iterations, ADAMW produced a relative $L_2$ error of 0.1770 and a final loss of $3.2537\times10^{0}$, outperforming the smaller network under the same optimizer.

\begin{figure}[H]
  \centering
  \begin{subfigure}[b]{0.45\textwidth}
    \centering
    \includegraphics[width=\textwidth]{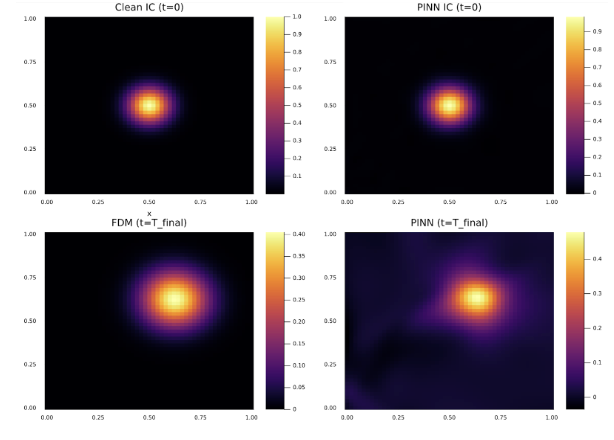}
    \caption{Solution plot}
    \label{fig:three:a}
  \end{subfigure}
  \hfill
  \begin{subfigure}[b]{0.45\textwidth}
    \centering
    \includegraphics[width=\textwidth]{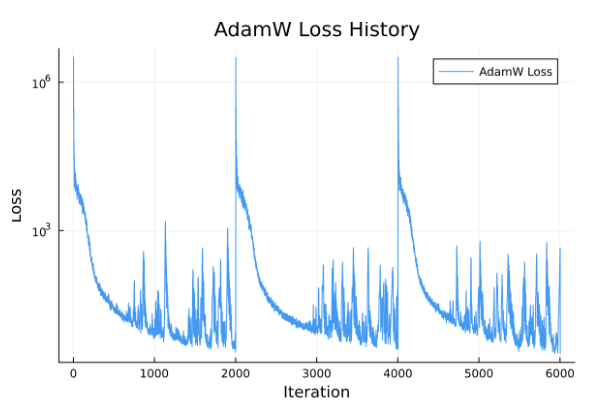}
    \caption{ADAMW Loss}
    \label{fig:three:b}
  \end{subfigure}
  \caption{Solution plots and AdamW loss history for Scenario 5 (9L-128N network, 6000 Adam iterations)}
  \label{fig:scenario5-adamw}
\end{figure}

\paragraph{ADAM with Varying Learning Rates:}  
\begin{itemize}
  \item \textbf{LR = 0.005, 6,000 iterations (Scenario 6):} Achieved an $L_2$ error of 0.1349 and the lowest final loss in this sweep, $2.5142\times10^{0}$.
  \begin{figure}[H]
  \centering
  \begin{subfigure}[b]{0.45\textwidth}
    \centering
    \includegraphics[width=\textwidth]{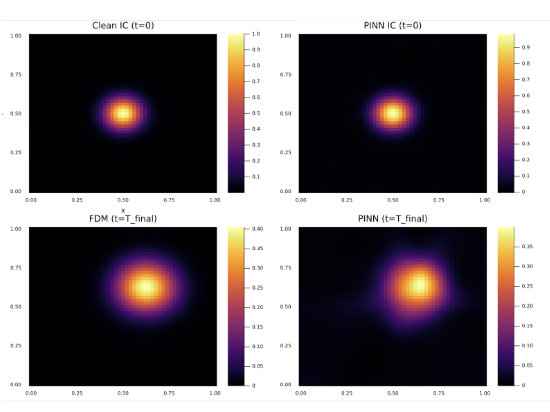}
    \caption{Solution plot}
    \label{fig:three:a}
  \end{subfigure}
  \hfill
  \begin{subfigure}[b]{0.45\textwidth}
    \centering
    \includegraphics[width=\textwidth]{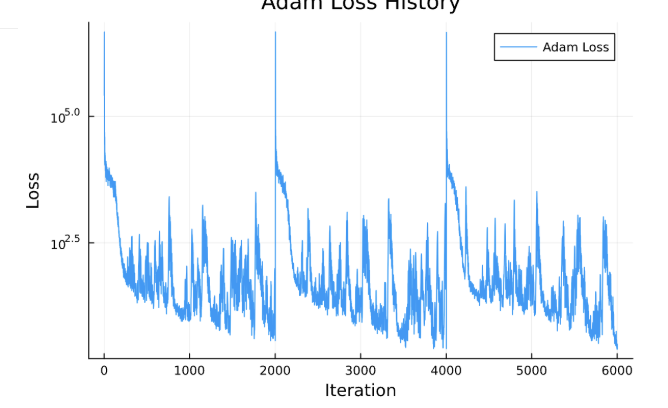}
    \caption{ADAM Loss}
    \label{fig:three:b}
  \end{subfigure}
  \caption{Solution plots and Adam loss history for Scenario 6 (9L-128N network, 6000 Adam iterations, 0.005 LR)}
  \label{fig:scenario5-adam-lbfgs}
\end{figure}

  \item \textbf{LR = 0.006, 6,000 iterations (Scenario 8):} Produced the best $L_2$ error overall, 0.08245, with a loss of $3.9296\times10^{0}$. The “PINN ($t=T_{\mathrm{final}}$)” plot (Figure~Z) shows a clean solution that closely matched the reference FDM.
  \begin{figure}[H]
  \centering
  \begin{subfigure}[b]{0.45\textwidth}
    \centering
    \includegraphics[width=\textwidth]{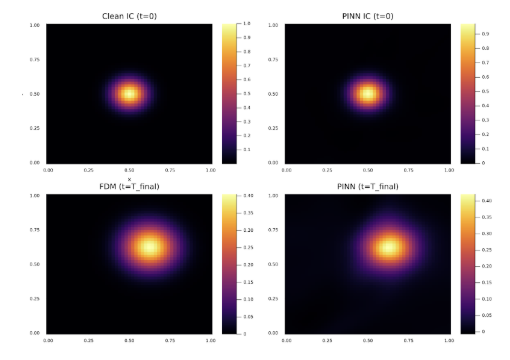}
    \caption{Solution plot}
    \label{fig:three:a}
  \end{subfigure}
  \hfill
  \begin{subfigure}[b]{0.45\textwidth}
    \centering
    \includegraphics[width=\textwidth]{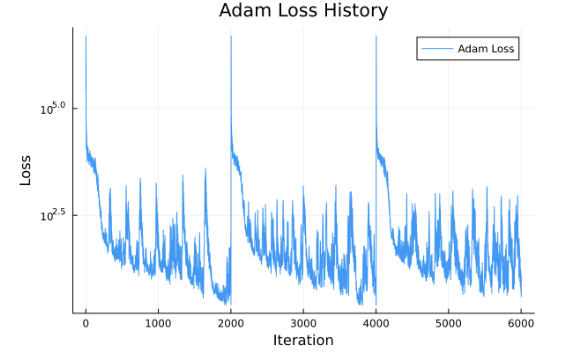}
    \caption{ADAM Loss}
    \label{fig:three:b}
  \end{subfigure}
  \caption{Solution plots and Adam loss history for Scenario 8 (9L-128N network, 6000 Adam iterations, 0.006 LR)}
  \label{fig:scenario8-adam}
\end{figure}

  \item \textbf{LR = 0.006, 10,000 iterations (Scenario 9):} Maintained an excellent $L_2$ error of 0.1294 and achieved the lowest overall final loss, $1.2614\times10^{0}$.
  \begin{figure}[H]
  \centering
  \begin{subfigure}[b]{0.45\textwidth}
    \centering
    \includegraphics[width=\textwidth]{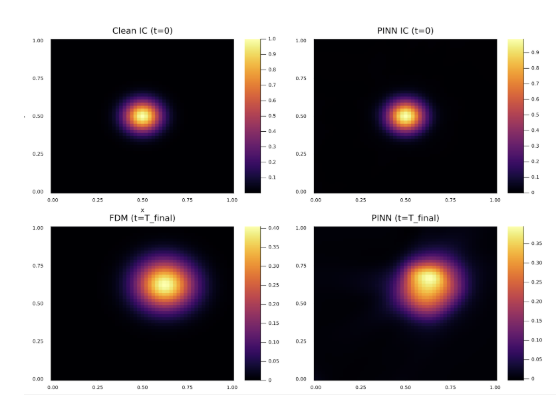}
    \caption{Solution plot}
    \label{fig:three:a}
  \end{subfigure}
  \hfill
  \begin{subfigure}[b]{0.45\textwidth}
    \centering
    \includegraphics[width=\textwidth]{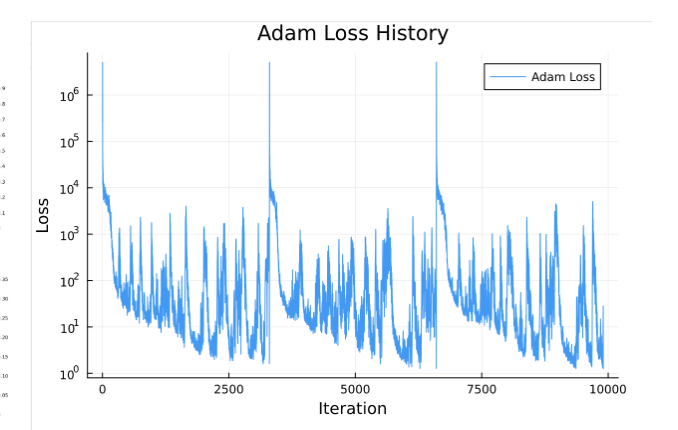}
    \caption{ADAM Loss}
    \label{fig:three:b}
  \end{subfigure}
  \caption{Solution plots and Adam loss history for Scenario 9 (9L-128N network, 10000 Adam iterations, 0.006 LR)}
  \label{fig:scenario9-adam}
\end{figure}

  \item \textbf{Other LRs:}  
    \begin{itemize}
      \item LR = 0.002, LR = 0.007, LR = 0.008, and LR = 0.01 resulted in higher $L_2$ errors and/or losses.  
      \item In particular, LR = 0.01 performed poorly compared to the midrange values.
    \end{itemize}
\end{itemize}

The visual results for the best runs (Scenarios 6, 8, and 9; Figures~a, b) show excellent initial condition learning and PINN solutions at $T_{\mathrm{final}}$ that qualitatively match the FDM data, with fewer artifacts than those of the smaller network.

\subsubsection{Performance with “9 Layers, 256 Neurons” Architecture}

This even larger architecture was tested using the ADAM optimizer with a learning rate of 0.006.

\paragraph{Scenario 10 (6,000 iterations)}  
Achieved a relative $L_2$ error of 0.22307 and a final loss of $5.3366\times10^{0}$ after 6,000 ADAM iterations. However, the training time was substantially longer (approx.\ 3,255.62s).

\begin{figure}[H]
  \centering
  \begin{subfigure}[b]{0.45\textwidth}
    \centering
    \includegraphics[width=\textwidth]{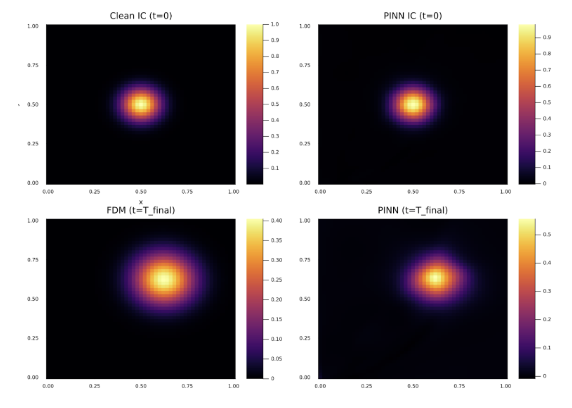}
    \caption{Solution plot}
    \label{fig:three:a}
  \end{subfigure}
  \hfill
  \begin{subfigure}[b]{0.45\textwidth}
    \centering
    \includegraphics[width=\textwidth]{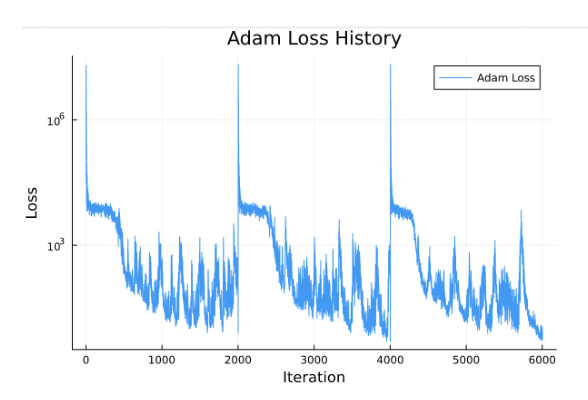}
    \caption{ADAMW Loss}
    \label{fig:three:b}
  \end{subfigure}
  \caption{Solution plots and Adam loss history for Scenario 10 (9L-256N network, 6000 Adam iterations, 0.006 LR)}
  \label{fig:scenario10-adam-lbfgs}
\end{figure}

Overall, the “9L, 256N” configuration did not outperform the “9L, 128N” architecture in error metrics, yet incurred much longer training times (e.g., $\sim$3,255s vs.\ $\sim$1,168s for Scenario 8). This suggests diminishing returns when scaling width without additional hyperparameter tuning or extended training.

\subsubsection{Loss Histories}

The loss history plots for the Adam and AdamW optimizers (e.g., Figure~L1 for Scenario~5, Figure~L2 for Scenario~8) typically exhibit a rapid decrease during the initial iterations, followed by a more gradual reduction or a noisy plateau. In several cases, the Adam/AdamW curves display large, sharp spikes where the loss temporarily increases significantly before recovering, reflecting the optimizer’s exploration of complex regions of the loss landscape.

When L-BFGS is used as a refiner (e.g., Scenario~1, Figure~LBFGS1), its loss history shows fewer iterations and often a smoother descent from the Adam endpoint. However, in these experiments, the L-BFGS refinement does not always yield a lower final loss compared to the best Adam/AdamW runs.

\subsubsection{Computational Efficiency}

The finite difference method (FDM) solver was consistently fast, requiring approximately 0.25–0.28 s to generate the $51\times51\times101$ reference dataset.  

PINN training times varied significantly with optimizer choice, network architecture, and iteration count:
\begin{itemize}
  \item The fastest training runs (e.g., ADAMW with 1,000 iterations in Scenario 8) were completed on the order of 50 s.
  \item The longest training time was approximately 3,255 s for ADAM with 6,000 iterations using the 9L–256N network (Scenario 10).
  \item The most accurate configurations, Scenario 8 (9L–128N, ADAM, 6,000 iterations) and Scenario 9 (9L–128N, ADAM, 10,000 iterations) required around 1,168 s and 1,987 s, respectively, illustrating a clear trade-off between accuracy and computational cost.
\end{itemize}

Once trained, PINN inference remained highly efficient, taking only about 0.024 s for a full field prediction at $T_{\mathrm{final}}$.

\subsubsection{Summary of Best Results}

The investigation reveals that Scenario 8 (9 Layers, 128 Neurons, ADAM optimizer, learning rate $0.006$, 6\,000 iterations) yielded the lowest relative $L_2$ error of approximately $8.25\%$. Close to this, Scenario 9 (9 Layers, 128 Neurons, ADAM optimizer, learning rate $0.006$, 10\,000 iterations) achieved a very good relative $L_2$ error of $12.94\%$ and the lowest overall final loss of $1.2614\times10^{0}$. These results underscore the significant impact of both network architecture and learning rate tuning on achieving high accuracy. The larger “9L, 128N” network generally outperformed the “9L, 64N” configuration, while the “9L, 256N” network did not offer further accuracy improvements and incurred substantially higher computational cost.

\section{Benchmarking: PINN vs. Finite Difference Method (FDM)}

To evaluate the efficacy of the Physics‐Informed Neural Network (PINN) as an alternative to traditional numerical solvers, we conducted a direct comparison against the Finite Difference Method (FDM) used to generate the reference data. The analysis focuses on two key aspects: solution accuracy and computational cost.

\subsection{Accuracy Comparison}

The primary metric for accuracy is the \emph{Relative \(L_2\) Error}, which quantifies the normalized difference between the PINN’s predicted solution \(u_{\mathrm{pred}}\) and the clean FDM reference solution \(u_{\mathrm{ref}}\) at the final time step \(T_{\mathrm{final}} = 0.25\). It is defined as:
\begin{equation}
\text{Relative }L_2\text{ Error}
= \frac{\|u_{\mathrm{pred}} - u_{\mathrm{ref}}\|_{2}}
       {\|u_{\mathrm{ref}}\|_{2}}.
\end{equation}

Across the various experiments, the optimal PINN configuration yielded impressive accuracy. The best result was achieved in Scenario 8, which utilized a “9L, 128N” architecture with the ADAM optimizer and a learning rate of 0.006. This model achieved a Relative \(L_2\) Error of approximately \(0.0825\) (8.25\%).

\subsection{Computational Cost Comparison}

The computational cost was evaluated by timing three distinct phases of the workflow using Julia’s \texttt{@BenchmarkTools.jl} package: FDM data generation, PINN training, and PINN inference. Table~\ref{tab:cost_comparison} summarizes the timings for a representative high‐performing scenario (Scenario~8).

\begin{table}[H]
  \centering
  \caption{Computational Cost Comparison: FDM vs.\ PINN (Scenario 8)}
  \label{tab:cost_comparison}
  \begin{tabular}{|l|l|l|p{6cm}|}
    \hline
    \textbf{Task} & \textbf{Method} & \textbf{Time (s)} & \textbf{Notes} \\
    \hline
    Data Generation / Single Solution 
      & FDM 
      & $\approx0.26$ 
      & Time to compute the entire $51\times51\times101$ solution grid. \\
    \hline
    Model Training 
      & PINN 
      & $\approx1168.53$ 
      & Time for 6\,000 ADAM iterations on the “9L, 128N” network (one‐time, offline cost). \\
    \hline
    Inference (Full Field Prediction) 
      & PINN 
      & $0.024$--$0.029$ 
      & Time for the trained PINN to predict the solution over the full $51\times51$ grid at one time step. \\
    \hline
  \end{tabular}
\end{table}

\section{Discussion and Conclusion}
The results from our experimental scenarios provide key insights into the effective application of Physics‐Informed Neural Networks for modeling the advection‐diffusion equation. The most striking finding is the critical role of network architecture and learning rate in achieving high accuracy. Our investigation revealed that a 9‐layer, 128‐neuron network significantly outperformed both a smaller (“9L, 64N”) and a wider (“9L, 256N”) architecture, suggesting a “sweet spot” for model capacity. The wider 256‐neuron network, despite its higher capacity, yielded worse accuracy and was computationally far more expensive, indicating that simply increasing network size can be counterproductive without further hyperparameter tuning, potentially due to a more challenging, non‐convex optimization landscape. Furthermore, we identified a narrow optimal range for the Adam learning rate around 0.005–0.006 for this best‐performing architecture, highlighting the sensitivity of PINN training to optimizer hyperparameters and underscoring the necessity of systematic tuning.

The training dynamics also offered valuable insights. The loss history plots for the Adam and AdamW optimizers consistently exhibited a rapid initial decrease followed by noisy plateaus with large, periodic spikes. These spikes likely represent the optimizer exploring different regions of the complex loss surface to escape local minima, a characteristic of adaptive first-order methods \cite{Schmidt2020}. While L‐BFGS provided a smoother descent, which is typical for a quasi-Newton method approaching a minimum \cite{Nocedal1980}, it was ineffective as a standalone optimizer and did not consistently improve upon a well‐converged Adam solution in these experiments, suggesting Adam’s exploratory nature was more beneficial for navigating the broader loss landscape of this specific problem, while L-BFGS struggled to find better minima without a good starting point, a known consideration for its use in PINNs \cite{Jarlebring2022}.

In summary, this paper successfully developed and benchmarked a PINN framework capable of solving the 2D advection‐diffusion equation for modeling ocean pollution. We demonstrated that a hybrid loss function, which strongly enforces the initial condition with a heavily weighted data‐driven term while also fitting to noisy reference data and respecting the underlying PDE, is critical for obtaining physically plausible and accurate solutions. Our systematic investigation culminated in identifying an optimal configuration—a 9‐layer, 128‐neuron network trained with the ADAM optimizer at a learning rate of 0.006—that achieved a relative $L_2$ error of approximately 8.25\% against a high‐resolution FDM solution. This highlights that PINN performance is highly sensitive to the interplay between network architecture and optimization hyperparameters.

While computationally intensive to train, the resulting PINN serves as a highly efficient surrogate model, capable of near‐instantaneous inference (approximately 0.024 s for a full‐field prediction), which is a key advantage over traditional solvers that must recompute the entire solution for any change in parameters. A key limitation of this study is the use of a constant velocity field, as real‐world ocean currents are spatially and temporally heterogeneous. However, the success of this framework with a constant field establishes a robust baseline for applying PINNs to environmental science and provides a clear methodology for hyperparameter exploration. Future work will focus on extending this framework to incorporate real‐world, non‐constant oceanographic data, moving closer to a practical tool for environmental monitoring and prediction.

\bibliographystyle{unsrtnat}

\end{document}